\documentclass{MITcsail}

\begin{document}

\title{\vspace{6mm}Counting Objects in a Robotic Hand 
(submitted to IROS, copyright may be transferred.)
}

\author{Francis Tsow, Tianze Chen, and Yu Sun
\thanks{Computer Science and Engineering Department, University of South Florida, Tampa, FL 33620, USA. Email: \texttt{\{tft,tianzechen,yusun}@usf.edu}}

\maketitle

\begin{abstract}
A robot performing multi-object grasping needs to sense the number of objects in the hand after grasping. The count plays an important role in determining the robot's next move and the outcome and efficiency of the whole pick-place process. This paper presents a data-driven contrastive learning-based counting classifier with a modified loss function as a simple and effective approach for object counting despite significant occlusion challenges caused by robotic fingers and objects. The model was validated against other models with three different common shapes (spheres, cylinders, and cubes) in simulation and in a real setup. The proposed contrastive learning-based counting approach achieved above 96\% accuracy for all three objects in the real setup. 

\end{abstract}

\section{Introduction}
\begin{figure}[h]
\graphicspath{ {./images1/} }
\centering
\includegraphics[scale=0.5, width=\columnwidth]{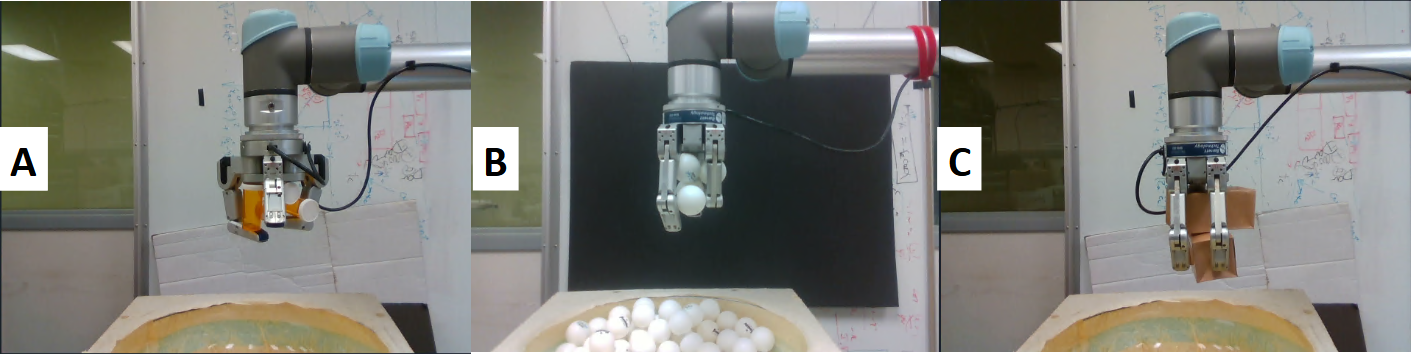}
\caption{One of the main challenges of predicting the number of objects in a robotic hand is occlusion. Other challenges include shadows, background, reflections, etc. The robotic hand had (A) 5 bottles, (B) 4 balls, and (C) 4 boxes in it.}
\label{challenge}
\end{figure}

Robots have the potential to perform many productive tasks, including in manufacturing, service industries, and logistics. One of the common actions required is picking up objects from one container and transferring them to another. Picking up objects from a bin is commonly referred to as a bin-picking problem in robotics. Usually, a robotic system picks one object at a time. If the picking is successful, the robot will either drop the object into the target bin or place it at a target location \cite{agrawal2010vision}. Determining the number of successfully picked and transferred objects that are unobscured \ref{challenge} or heavy may not be too challenging due to the availability of pressure sensors or cameras. However, to accurately count the number of objects while it is being picked up and transferred by a robotic hand is a challenging research problem. 

To determine the number of objects grasped in a robotic hand, in addition to the typical challenges faced by other object counting applications, the robotic hand is usually close to the targeted objects and often results in partial obscurity of the objects by the robotic hand. Moreover, the objects tend to be packed in the grasp of the robotic hand, further leading to the obscurity of the objects, with the difference being that the occluding objects are some of the targeted objects themselves. Thirdly, due to the lightweight of the targeted objects, the limited sensitivity of the tactile sensors, and the relatively heavy weight of the robotic hand, using the force sensor at the wrist of the robotic hand to estimate the number of objects by their additional weights compared to the object-free weight of the robotic hand is also difficult. For instance, the wrist sensor on a UR5e has a resolution of about two newtons. This means that the object will have to be around 200 grams for it to be readily detectable. The weight of a typical sugar cube is merely a couple of grams. As a result, we need to come up with an alternative way to estimate the number of objects grasped in a robotic hand for practical real-world applications.

When a robotic system is designed to pick multiple objects at once for efficiency, multi-object grasping (MOG) technologies are applied \cite{Chen2021,Sun2022, DBLP:journals/corr/abs-2112-09829, Sun2022a}. MOG approaches give robots the capability to pick multiple objects and drop them into another container. It saves on the number of picking and transferring \cite{shenoy2021multi}. So, MOG is preferable to single object grasping for its superior efficiency. 

Picking and transferring multiple objects on demand is a challenging problem. Our previous work assumed prior knowledge of the number of objects and demonstrated a set of novel strategies to efficiently grasp multiple objects from a bin and transfer them to another \cite{shenoy2021multi}. In our most recent work, the enhanced MOG strategy significantly boosts efficiency in object transfer tasks by up to 60\%. However, knowing the number of objects in the hand before dropping is both critical for the approach to work and challenging. If the robot over-estimated the count in the hand, the robot would have to pick the missing objects again, thus, reducing efficiency. On the other hand, for some situations, a re-do is not possible. For instance, if a robot is supposed to pick three sugar cubes and drop them into a cup of coffee but instead pick and drop four into the coffee, it is impractical to extract the extra sugar cube from the coffee as the cubes would have dissolved.

The focus of this paper is to identify a simple and effective approach to estimate the number of objects grasped in the robotic hand after the hand has been lifted from the source container.

The contributions of this paper include 1) the development of a modified loss function for the supervised contrastive learning-based classifier as a simple and effective approach for object-count prediction, and 2) validating the developed model against other popular object counting models in three simulated and real objects (sphere, cylinder, and cube).   

\section{Related works}
Object counting is to estimate the number of objects in a given setting, be it an image, video, or part of an image or video. It has found many applications, including those in industries, security, agriculture, and research \cite{9399607, s20072145, Lawal2021, rs13091619, s21144803, Hobbs2021, Sadrfaridpour2021, 7780439, Heinrich2020, doi:10.1080/21681163.2016.1149104}. While conventional object detection approaches relied on segmentation, recently, deep learning techniques, such as various variants of the convolutional neural network, had been applied \cite{Sadrfaridpour2021} to object counting and provided impressive results even when the objects were not easily distinguishable. However, conventional deep learning approaches require a large training data set. Generally speaking, there are the detection and classification approaches to this task. The detection approach typically localizes the objects (such as object detection, where object count is predicted by counting the number of identified objects) while the classification approach identifies features in the image for classifying them into classes of a number of objects. There is also the regression approach that maps features in the image to numbers (such as density maps where the sum of the density maps produces the object counts) or classes. 

\subsection{Object detection techniques}
Several groups applied various machine learning techniques to the object localization problem that allowed learning from the data without precise modeling of the objects or the environment. Once the objects have been localized (e.g., using bounding boxes), the number of counting boxes can be counted to yield the object count. Support vector machine and other techniques have been applied to solve object localization problems \cite{Brahmbhatt2015, Liu2018}. The model identified bounding boxes and allowed obscured pixels to be properly assigned to the correct objects using assignments of its neighboring pixels. Deep learning techniques can further enable autonomous learning and reduce the need of hard-coded features \cite{Lenz2015, Nagata2021}. Approaches such as reinforcement learning \cite{Caicedo2015}, convolutional neural networks, and their variants, including region-based convolutional neural network (R-CNN), faster R-CNN, and single-shot deep learning techniques, have been employed for localization. In addition, single-shot techniques such as You Only Look Once (YOLO) \cite{9399607, s20072145, Lawal2021, rs13091619, s21144803} had been applied to obtain bounding boxes for the objects. The bounding boxes identified objects can then be counted \cite{s21144803} to yield an object count. 

\subsection{Classification/Regression techniques}
Contrastive learning had been developed and applied to image classification \cite {Chen2020, Khosla2020}. This approach is simple and intuitive and has been demonstrated to result in better accuracy with less data and less detailed annotations as compared to other counting techniques. Transformers have also been used in object detection and image classification and have been expanded to object localization applications \cite{Carion2020}. The advantages of using transformers include conceptual simplicity as compared to other complicated approaches such as YOLO, attention among input patches, and parallel input patch processing. Carion et al. combined convolutional neural network with transformers encoder-decoder and demonstrated that the use of transformers could remove the need for components that capture prior knowledge, e.g., non-max suppression or anchors, and allowed end-to-end training and all objects to be predicted in parallel \cite{Carion2020}. Dosovitskiy et al. split the input image up into position-embedded patches to take advantage of spatial relationships among the different patches and fed them into the transformer model in parallel, demonstrating the use of transformers in image recognition \cite{Dosovitskiy2020}.

Image density used in object counting had been particularly useful in crowd counting where many and/or overlapping objects were involved \cite{Lempitsky10b}. It avoided having to detect and localize objects \cite{8578218} but only to learn features of the type of object that was being counted. More recently, counting with density maps has expanded to object agnostic and few-shot counting techniques to address the needs of large training data sets \cite{Yang, Hobley2022, Ranjan2021, Shi2022}. Although this approach could also potentially address occlusion issues, based on our experiment, the density map approach worked better in handling self-occlusion than with occlusion by different types of objects.

\section{Methodology}

To identify a simple and effective approach to estimate the number of objects grasped in the robotic hand after the hand has been lifted from the source container, we proposed a simple hand-eye system with a camera at the side of the bin looking at the robotic hand while it is lifted from the bin, together with an object counting model.

\begin{figure}[h]
\graphicspath{ {./images1/} }
\centering
\includegraphics[scale=0.5, width=\columnwidth]{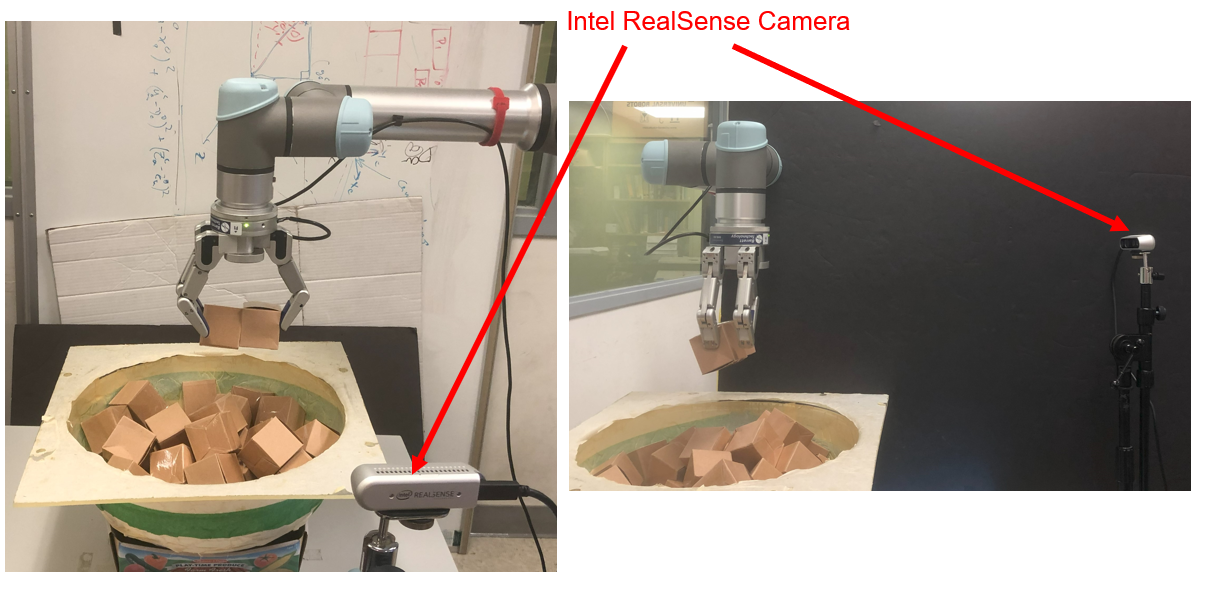}
\caption{Front and side views of the experimental setup with the robotic arm and camera.}
\label{setup1}
\end{figure}

\subsection{System setup}
Figure \ref{setup1}(B) shows an example of the setup with an UR5e, a BarrettHand\textsuperscript{\texttrademark}, and an  Intel\textsuperscript{\textregistered} Realsense\textsuperscript{\texttrademark} camera. To overcome occlusion, the robotic arm rotated the hand through $270^{\circ}$ to allow the camera to observe the hand and objects in the hand from four different perspectives (back, front, left, and right). 

To avoid bias due to imbalanced classes, we balanced the number of images for different numbers of objects in the data set for the classification approach during training. We repeated higher object-count images sequentially and looped over the entire set with the same object count, which typically were rarer than those of the lower object-count set so that different object counts had the same number of images. It is noted that we did not balance the data set for the object detection approach. It is also noted that the test data set was not balanced. We split the input data into batches of 32, which were empirically determined to optimize for memory, efficiency, and performance. 

\begin{figure}[h]
\graphicspath{ {./images1/} }
\centering
\includegraphics[scale=0.5, width=\columnwidth]{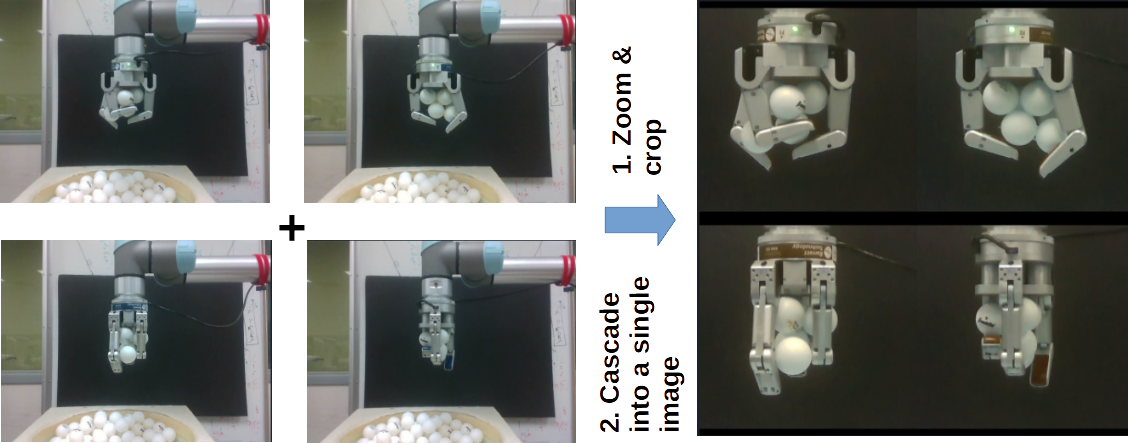}
\caption{Overview of the preprocessing of the input images.}
\label{multi-view}
\end{figure}

\subsection{Contrastive Counting}
The most pressing challenge of counting objects in a robotic hand is obtaining data in the real system. Getting successful grasps in real systems, in general, requires a significant amount of time. Usually, data sets of robotic grasping in real systems are small. Multi-object grasping exaggerates this challenge with two factors \cite{Chen2021}. First, it is significantly more challenging to get successful grasps of multiple objects in real systems. Secondly, it is much more challenging to obtain samples of a robotic hand holding many objects. Therefore, the grasping data sets of multi-object grasping in real systems are small and unbalanced. To address these challenges, we designed a contrastive counting approach that is based on supervised contrastive learning. In our approach, we trained a supervised contrastive feature vector using supervised contrastive learning first and then built a classifier on the trained, supervised contrastive feature vector. 

\begin{figure}[h]
\graphicspath{ {./images1/} }
\centering
\includegraphics[scale=0.5, width=\columnwidth]{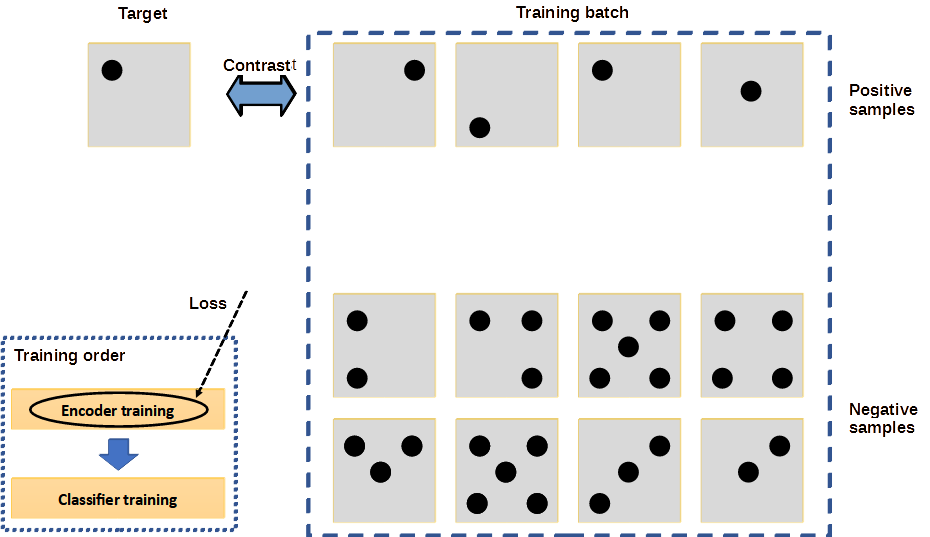}
\caption{Overview of contrastive learning training.}
\label{contrastive1}
\end{figure}

\paragraph{Contrastive feature vector}
We use ResNet-50 architecture to design an encoder as in \cite{Khosla2020}. It outputs a 2048-dimensional feature. 
The encoder is trained with a projection head to learn embedding features about the inputs to minimize loss. Assume a robot hand can hold $n$ or less objects. When we trained the supervised contrastive feature vector, we grouped the images of holding $i$ objects as positive samples and the rest (holding $1$, ..., $i-1$, $i+1$, ... $n$) as negative samples. Then we did this iterative with $i$ rotated from $1$ to $n$ again and again till it converged. Figure \ref{contrastive1} shows one example for $i=1$ when it is implemented for mini-batch stochastic gradient descent training. Training for the encoder was performed with the Adam optimizer, a learning rate of 0.001, and with the modified supervised contractive learning loss (described below). The conventional supervised contrastive learning loss can be written as \cite {Khosla2020}:

\begin{equation}\label{eq1}
Loss = 
\sum_{i \in I} \frac{-1}{|P(i)|} \sum_{p \in P(i)} log \frac{exp(z_i \cdot \frac{z_p}{\tau})}{\sum_{\alpha \in A(i)} exp(z_i \cdot \frac{z_a}{\tau})} \
\end{equation}
where z is the output after the projection layer, i represents the index of the input sample in the input batch, j represents the index of the positive sample i is compared to, \( \tau \) is the temperature parameter, and A(i) is the set of negative samples. P(i) is the set of indices of positive samples and \( |P(i)| \) is the number of samples in the set.

\paragraph{Classifier using contrastive feature vector} 
The classifier comprises dense layers (with dropout), with the last layer being a softmax layer. 
Without the projection head, the pre-trained encoder was used to train the classifier to output the classes of different numbers of objects in the robotic hand.
Training for the classifier was performed with the Adam optimizer, a learning rate of 0.001, and with the cross entropy loss.

\paragraph{Improvement with Fourier transform loss} 
Taking advantage of the repeating patterns in the image when multiple objects were grasped by the robotic hand, we added a Fourier transform component in the loss function as shown in Equation \ref{eq3} and \ref{eq4}. In our experiments, we noticed that by adding the Fourier transform component to the loss function, the count prediction accuracy of some of the higher object-count was slightly better than or similar to those using the conventional supervised similarity loss function depending on the shapes and sizes of the test objects, and the quality of the inputs. 

\begin{equation}\label{eq3}
z' = 
w0 * z + w1 * \mathcal{F}(z) \
\end{equation}

\begin{equation}\label{eq4}
Loss = 
\sum_{i \in I} \frac{-1}{|P(i)|} \sum_{p \in P(i)} log \frac{exp(z'_i \cdot \frac{z'_p}{\tau})}{\sum_{\alpha \in A(i)} exp(z'_i \cdot \frac{z'_a}{\tau})} \
\end{equation}
where z' is the weighted (w0 and w1) cascaded sum of z and a fast Fourier Transform version of z.

\subsection{Other data-driven approaches}
To develop a model that allows us to accurately determine the number of objects grasped by a robotic hand, we adopted other object-counting approaches based on classification and object detection. 

Object detection approaches have the potential advantage that 1) it is less sensitive to imbalanced data, and 2) it may be more robust for transfer learning (i.e., more tolerant to background, color, similar shapes, and sizes). Objection detection learns to identify a specific object in an image based on its features. As a result, the influence of the background can be mitigated. Similarly, if the identifying features learned are generalized enough for the class of objects (i.e., key features for positive identification of the object class), the prediction can be more tolerant against differences such as small differences in shapes, sizes, and colors. Therefore, we explored the use of object detection in our application. We applied the object detection algorithm based on the popular YOLO technique proposed by Redmon et al.\cite{Redmon2016}. Our approach took images from four different views and combined them into a single image \ref{multi-view}. The image was subsequently divided into grid cells and predictions were made of centers, widths, and heights of the bounding boxes across different scales in addition to the confidences and classes for the detected objects. 


Transformers can incorporate information from the different camera perspectives by cascading images from the four different camera perspectives as a single input image (same as what we did in the object detection approach). The combined image was then divided into smaller patches (12x12 pixels), which were then fed into dense layers and projected into a dimension of 128. 
The transformer block used was that of the typical transformer \cite{Vaswani2017}.
Since a large estimation error on the number of objects is inefficient and can have an irreversible undesirable effect, we implemented a loss function that is sensitive to the estimation error (equation \ref{eq2}), where {$T_k$} and {$P_k$} are the ground truth and predicted number of objects, {$p_k$} is the probability for the predicted number of objects. 

\begin{equation}\label{eq2}
Loss = [(1/70) * (T_k - P_k) - 1 ] * log(p_k + 10^{-20})
\end{equation}

The density-based approach has been demonstrated in the crowd and agricultural product counting; we applied CSRNet \cite{8578218} to our problem. We followed the model developed by Li et al. \cite{8578218} with some adjustments. 


However, in our experiment with the 40mm-balls, although our model was able to learn to predict density maps representing the objects in the robotic hand, the results were significantly worse than the other approaches. Our result appeared to agree with other groups' work wherein the density-based approach introduced noise to the model, probably due to the use of the pre-processing kernels to generate density maps. 
As a result, we abandoned this approach in when processing other shapes in other studies.

Lastly, we have already explored using pre-trained models to reduce the need for a large data set.
We selected several pre-trained large vision transformer-based models released by Google (google/vit-base-patch16-224-in21k) \cite{wu2020visual}, Facebook (Facebook/dinov2-base) \cite{oquab2023dinov2}, and Microsoft (Microsoft/swin-tiny-patch4-window7-224) \cite{DBLP:journals/corr/abs-2103-14030} available on Hugging Face. These were pre-trained with various image collections (e.g., ImageNet-1K, ImageNet-21k, ImageNet-22k, Google Landmarks v2, etc.). We first tested the pre-trained models without fine-tuning but obtained unsatisfactory results. Subsequently, we fine-tuned them with our data set. We fine-tuned the pre-trained models with simulation images and then tested them with unseen test simulation images. For real images, we first fine-tuned the pre-trained models with simulation images. Then, we fine-tuned again with real images as the number of real images available was limited. 


\section{Evaluation and Results}
\subsection{Data collection}
To train and evaluate the models, we collected data in both the simulation and the real setup. The simulation setup was a replication of the real setup except for the camera. To speed up the data collection, instead of letting the robotic arm rotate the hand, the setup in the simulation had four hand-facing cameras at the same height and at the four locations to the back, front, left, and right of the hand. 

\begin{figure}[h]
\graphicspath{ {./images1/} }
\centering
\includegraphics[scale=0.5, width=\columnwidth]{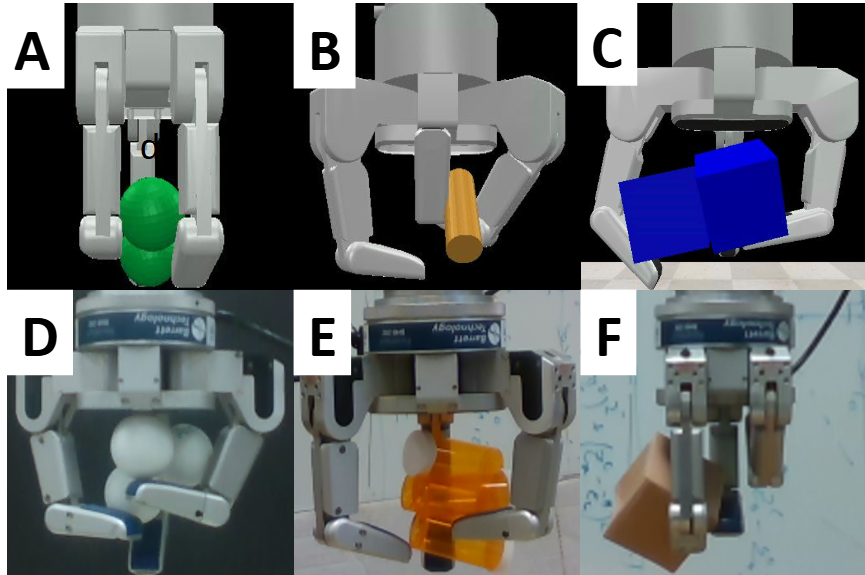}
\caption{(A) Simulated robotic hand holding spheres. (B) Simulated  robotic hand holding cylinders. (C) Simulated  robotic hand holding cubes. (D) Real robotic hand holding ping-pong balls. (E) Real  robotic hand holding pill bottles. (F) Real robotic hand holding cubes.}
\label{objects1}
\end{figure}

\paragraph{Objects}
To evaluate the approaches, we selected three common shapes in logistics: spheres, cylinders, and cubes. 
The detailed specifications of the objects in the simulation and real setups are listed in Table \ref{table-shapes}. All the simulated objects are of randomized color generated by randomly picking a number between 0 and 255 among all three channels. The real objects used in this study included 40mm-diameter ping-pong balls (spheres), pill bottles (cylinders), and gift boxes (cubes). Figure \ref{objects1}(A-F) shows several examples of the objects. 

\begin{table}
\caption{\label1 Summary of objects used in experiments.}
\begin{center}
\begin{tabular}{ |p{8em}|p{8em}|p{4em}|} 
 \hline
  Object label & Dimension & Mass \\ 
 \hline
 \multirow{1}{10em}{Ping-pong ball} & diameter=40mm & 2.7g \\ 
 \hline
 \multirow{1}{5em}{Sphere} & d=40mm & 2.7g \\ 
 \hline
 \multirow{1}{5em}{Pill bottle} & d=26mm, h=63mm & 7g\\
 \hline
 \multirow{1}{10em}{Cylinder} & d=23mm, h=61mm & 6.6g \\ 
 \hline
 \multirow{1}{10em}{Gift box} & side=2inches & 5g\\
 \hline
 \multirow{1}{10em}{Cube} & side=2inches & 5g \\ 
 \hline
\end{tabular}
\end{center}
\label{table-shapes}
\end{table}

\paragraph{Data Collection} To collect data for training and testing, we performed MOG routines as described in \cite{Chen2021, shenoy2021multi}.

\begin{table}
\caption{Multi-view data set sizes of various objects used in experiments.}
\begin{center}
\begin{tabular}{ |p{12em}|p{2em}|p{2em}|p{2em}|p{2em}|p{2em}|} 
 \hline
Object-Count & Zero & One & Two & Three & Four+ \\
 \hline
Ping-pong ball with bounding-box annotations & NA & 106 & 104 & 52 & 9 \\ 
 \hline
Ping-pong ball (all) & 303 & 428 & 348 & 140 & 31 \\ 
 \hline
Sphere & 1,211 & 2,674 & 1,752 & 558 & 105 \\ 
 \hline
Pill bottle & 60 & 60 & 40 & 25 & 15 \\
 \hline
Cylinder & 5,008 & 3,777 & 1,461 & 380 & 173 \\
 \hline
Gift box & 77 & 46 & 42 & 29 & 15 \\
 \hline
Cube & 1,221 & 381 & 140 & 40 & 18\\ 
 \hline
\end{tabular}
\end{center}
 \label{table-data}
\end{table}

\paragraph{Data set}
Table \ref{table-data} summarized the collected data of the different objects used in the experiments. The columns provided the breakdown of sample sizes of different counts. The numbers reflected the unbalanced nature of the data since the chances of grasping many objects in one grasp were much lower than grasping one or two of them. We noticed that for ping-pong balls, bounding boxes annotation did not apply for zero ball trials as there were no objects to annotate and is therefore indicated as NA (Not Applicable).
The collected data was divided for training and testing in a $9:1$ ratio. In the simulation, both the object counts and their bounding boxes were obtained automatically. For the real ping-pong balls, they were manually annotated. Since providing bounding boxes was much more costly, the data sizes for the object-detection-based approach in the real setup were much smaller (271 in total).

\subsection{Results and Analysis}


\subsubsection{Preliminary evaluation on 40mm simulated spheres}

We evaluated the supervised contrastive-learning-based classifier, object detection, transformer-based classifier, and fine-tuned pre-trained vision transformer models on the 40mm simulated spheres data set. 

Accuracies, as defined by the correct predictions over total cases of the said number of objects, were calculated for the various number of objects.
An overall accuracy was also calculated for a quick comparison of performance. In addition, not only were accuracy important to us, we also wanted to minimize the size of the count prediction error. This was especially so if the accuracy was not high. In other words, even if we failed to predict the actual number of objects, we preferred a prediction that was closer to the actual number to reduce the resulting penalty due to the wrong prediction. To this end, the root-mean-square errors (RMSE) between the predicted number of spheres and the actual number of spheres were calculated. The accuracy and RMSE are summarized in Tables \ref{table_acc_ob} and \ref{table_err_ob}, respectively.

\begin{table}
\caption{Summary of multi-view approaches prediction accuracy on simulated spheres.}
\begin{center}
\begin{tabular}{ |p{10em}|p{7em}|} 
 \hline
  Approach & Accuracy (\%) \\ 
 \hline
 \multirow{1}{13em}{Contrastive} & 92 \\
 \hline
 \multirow{1}{13em}{Object detection} & 96 \\ 
 \hline
 \multirow{1}{13em}{Transformer} & 82 \\
 \hline
 \multirow{1}{13em}{SWIN} & 97 \\
 \hline
 \multirow{1}{13em}{Google} & 96 \\ 
 \hline
 \multirow{1}{13em}{Facebook} & 97 \\ 
 \hline
\end{tabular}
\end{center}
\label{table_acc_ob}
\end{table}

\begin{table}
\caption{Summary of multi-view approaches RMSE on simulated spheres.}
\begin{center}
\begin{tabular}{ |p{13em}|p{3.5em}|} 
 \hline
  Approach & RMSE \\ 
 \hline
 \multirow{1}{13em}{Contrastive} & 0.29 \\
 \hline
 \multirow{1}{13em}{Object detection} & 0.22 \\ 
 \hline
 \multirow{1}{13em}{Transformer} & 0.45 \\
 \hline
\multirow{1}{13em}{SWIN} & 0.16 \\
 \hline
\end{tabular}
\end{center}
\label{table_err_ob}
\end{table}

\begin{figure}[h]
\graphicspath{ {./images1/} }
\centering
\includegraphics[scale=0.5, width=\columnwidth]{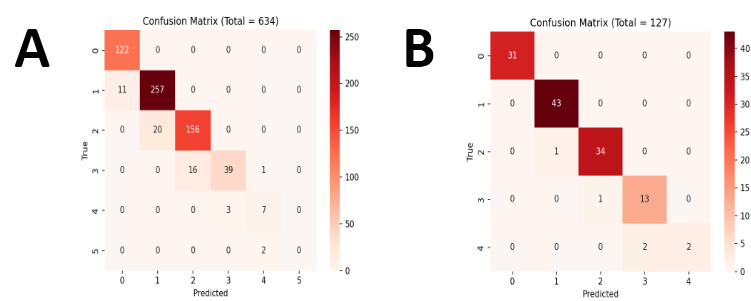}
\caption{Confusion matrices of supervised contrastive learning model with (A) simulated 40mm spheres, and (B) ping-pong balls.}
\label{contrast_sim_real_sphere}
\end{figure}

Figure \ref{contrast_sim_real_sphere} (A) shows the counting confusion matrix for the simulated 40mm spheres. Although it did not have the highest accuracy, it was still better than 90\% and had better accuracy with real objects (please see later section). It is noted that these models were trained from scratch, and the exact number of test images was slightly different depending on the exact splitting of the total image data set during the experiments. As for the trained and fine-tuned models, from Table \ref{table_acc_ob}, we can see that they were even more accurate with their count predictions. It is also noted that the training time for the pre-trained model was significantly shorter than training the naive models.


\subsubsection{Evaluation with other shapes}
Besides spheres, we used other shapes, including cylinders and cubes (Table \ref{table-shapes}). We applied the supervised contrastive learning-based classifier, object detection, transformer-based classifier, and pre-trained SWIN models to the simulation cylinders. The resulting accuracy and RMSE are summarized in Tables \ref{table_acc_cyl} and \ref{table_err_cyl}, respectively. 

As can be seen from the summarizing tables (Tables \ref{table_acc_cyl} and \ref{table_err_cyl}), the accuracy for the cylinders was lower than that of the spheres. This could be a result of the slender bodies and high length-to-width aspect ratio, making them easier to hide and harder to identify. Nevertheless, among the different approaches that we tried, the pre-trained model (SWIN) was the best and provided similar performance compared to the spheres. Since the SWIN model provided the best result, we applied it to the simulated cubes (width=2 inches), and an accuracy of 100\% was obtained.

It is also noted that the accuracy of the cubes was higher than that of the other shapes. This could be due to the relatively large size of the cubes (2 inches on each side), resulting in fewer objects being grasped together and reducing the chance of an object being completely or mostly obscured.

\begin{table}
\caption{Summary of multi-view approaches prediction accuracy on simulated spheres, cylinders, and cubes.}
\begin{center}
\begin{tabular}{ |p{7em}|p{5em}|p{5em}|p{5em}|} 
 \hline
  Approach & Sphere & Cylinder & Cube \\ 
 \hline
 \multirow{1}{13em}{Contrastive} & 92\% & 88\% & 97\% \\
 \hline
 \multirow{1}{13em}{Object detection} & 96\% & 60\% & 85\% \\ 
 \hline
 \multirow{1}{13em}{Transformer} & 82\% & 85\% & 97\% \\
 \hline
 \multirow{1}{13em}{SWIN} & 97\% & 97\% & 100\% \\
 \hline
\end{tabular}
\end{center}
\label{table_acc_cyl}
\end{table}

\begin{table}
\caption{Summary of multi-view approaches RMSE on simulated spheres, cylinders, and cubes.}
\begin{center}
\begin{tabular}{ |p{8em}|p{3.5em}|p{3.5em}|p{3.5em}|} 
 \hline
  Approach & Sphere & Cylinder & Cube\\ 
 \hline
 \multirow{1}{13em}{Contrastive} & 0.29 & 0.38 & 0.17 \\
 \hline
 \multirow{1}{13em}{Object detection} & 0.22 & 0.64 & 0.7 \\ 
 \hline
 \multirow{1}{13em}{Transformer} & 0.45 & 0.70 & 0.18 \\
 \hline
\end{tabular}
\end{center}
\label{table_err_cyl}
\end{table}

\subsubsection{Evaluation with real objects}

In real-world applications, the background, lighting, shadows, and other environmental factors can impact the quality of the images, resulting in poorer prediction outcomes. As a result, we tested the models with real objects (ping-pong balls, pill bottles, gift boxes) in real experiments with a physical robotic arm/hand and cameras. In this part of the study, we applied the transformer-based classification model, the supervised contrastive learning-based classification model (Figure \ref{contrast_sim_real_sphere} (B)), and the pre-trained Microsoft SWIN model that was first fine-tuned with simulation images and then with the limited real images. The object detection model for real objects was not applied as the performances of the other models were better with simulated objects, and getting bounding boxes for real objects was proven to be very laborious. We tested ping-pong balls, pill bottles, and gift boxes representing spheres, cylinders, and cubes, respectively. The results (accuracy and RMSE) were summarized in Table \ref{table_real_acc}.

\begin{table}
\caption{Summary of multi-view object-detection-based approach prediction accuracy (rms) on real objects.}
\begin{center}
\begin{tabular}{ |p{8em}|p{6em}|p{6em}|p{6em}|} 
 \hline
  Approach & Ping-pong ball & Pill bottle & Gift box \\ 
 \hline
 \multirow{1}{13em}{Contrastive} & 97\% (0.18) & 96\% (0.21) & 100\% (0.00)\\
 \hline
 \multirow{1}{13em}{Transformer} & 92\% (0.28) & 74\% (0.51) & 96\% (0.21)\\
 \hline
 \multirow{1}{13em}{SWIN} & 86\% (0.37) & 76\% (0.49) & 90\% (0.31)\\
 \hline
\end{tabular}
\end{center}
\label{table_real_acc}
\end{table}

From Table \ref{table_real_acc}, it can be seen that the accuracy of the gift boxes was generally better, similar to the simulation results. This was probably due to the larger size of the gift boxes as in the simulation experiments. Out of the three models, the supervised contrastive learning model provided the best prediction accuracy. This was different from the simulation model, where the supervised contrastive learning approach did not produce the best accuracy for the cylinders. One possible reason could be that the contrast of the pill bottles was better than that of the cylinders in the simulation. 

\section{Discussions and Conclusion}
\subsubsection{Proposed supervised contrastive learning model}
Contrastive learning was a good approach for this application as we had a significant fraction of negative samples to improve discrimination between signal and noise. In addition, conventional supervised contrastive learning loss enables larger contributions from larger contrast samples. Furthermore, conventional supervised contrastive learning loss allows more robust clustering of samples from the same number of objects \cite{Khosla2020}.

The proposed supervised contrastive learning model performed well with all the simulated and real objects. Importantly, it outperformed the other models with real images, which were more challenging due to background noise, varying lighting, shadows, etc. Although the pre-trained model performed better than the supervised contrastive learning model with simulated images, it did not perform as well with real images. We hypothesized that the pre-trained model required more training data for fine-tuning, which we only had limited sets as they were laborious to obtain. This highlights the simplicity and effectiveness of the supervised contrastive learning model. Furthermore, we added a term to the supervised contrastive loss in an attempt to capture frequency components in the image after feature extraction. It is noted that the modified loss function was very helpful in improving accuracy when there was not enough data (as was the case with tactile sensor readings only as inputs) and had the tendency to improve high object-count accuracy when images were used as inputs (probably due to its frequency sensitive nature). We used the supervised contrastive learning model and obtained the best accuracy among the different models, including the transformer, objection detection, and pre-trained vision transformer models.

\paragraph{Classification compared to object detection approach}
A classification counting approach imposes an upper-bound prediction based on the maximum number of objects on which the model was trained. On the other hand, an object detection approach predicts count based on the number of the identified objects. As a result, the predicted number of objects can exceed those of the training data. This has been demonstrated in experiments where the predicted maximum count will not exceed the number of classes the model was trained on. In lower object-count cases where the objects can be clearly seen and therefore identified, the object detection approach worked very well. However, the classification approach did not require specifying the exact location of the object, making it easier to prepare the ground truth for model training. 

\paragraph{Use of multi-view and tactile sensor as inputs}
We considered the availability of multiple perspectives (i.e., being able to look at the robotic hand from different angles) without cascading the different views into a single image, i.e., by taking the maximum prediction on the number of objects out of the four views (back, front, left, and right) for each sphere-grasping trial. The accuracy was higher when different views were combined and used as a single input. This suggested that the relationships among the different views provided useful information for the model's training.

\paragraph{Tactile Sensors} The Barrett Hand has three patches of tactile sensors on the fingertips and one patch in the palm. It can potentially help alleviate the occlusion challenges posed by the camera. We explored including tactile sensor readings as inputs to the various models. However, we found that tactile sensor readings did not typically help with prediction accuracy. On the other hand, when there were sufficient input images to the models, the tactile sensor readings appeared to confuse the model, resulting in lower prediction accuracy. This might be due to noise in the tactile sensor readings as our objects were light (thus low signals), while the tactile sensor had a relatively high noise compared to the signals. It is noted that using only tactile sensors as inputs to the models (without input images) allowed the model to predict. For instance, with the supervised contrastive learning model, with only the tactile sensor readings as inputs, the prediction accuracy was 64\% with an RMSE of 0.72. Although the accuracy was lower than the 92\% we obtained using the input images, the accuracy was higher than that of a random guess.

\paragraph{Benefits of pre-trained models} 
Finally, we leveraged pre-trained models that were released by groups that trained them with millions of images. We only needed a fraction of the time to fine-tune the models to our applications, and the resulting models outperformed the naive models that we only had limited data to train and were more susceptible to over-fitting and biases due to the lack of diversity in our data. In our experiments, the pre-trained models outperformed all of our naive models with simulated objects, although they did not produce the best accuracy with real objects. It is also noted that pre-trained models may also limit our ability to customize the architectures of the models.

In summary, we applied multi-view deep learning object detection and classification approaches to the problem of object counting in multi-object grasping in a robotic hand. We used both simulation images and real images in our experiments. We demonstrated in both cases that the models can learn from cascaded images from different camera perspectives and from multi-view relationships. We also proposed adding a Fourier transform component to the supervised contrastive learning loss function to aid in predicting object counts for multiple repeating objects and demonstrated its applications in our experiments. Further, while the classification approach may be easier to annotate and can accurately predict counts that it had been trained on (e.g., 0-4 spheres), it cannot predict unseen cases (e.g., 5-spheres and above). Moreover, from our experiments, the larger objects (cubes) appeared to make it easier for the models to predict the object counts accurately. It appeared that the larger objects might have reduced the occlusion problem as they were less likely to be completely blocked from viewing. 


\bibliographystyle{IEEEtran}

\bibliography{references}

\end{document}